\documentclass{article}
\usepackage{arxiv}

\usepackage{stmaryrd}

\usepackage{amsmath,amsfonts,amssymb}
\usepackage{mathrsfs}
\newcommand{\X}{{\mathbb X}}
\DeclareMathOperator{\pers}{pers}
\DeclareMathOperator{\Dgm}{Dgm}
\DeclareMathOperator{\St}{St}
\DeclareMathOperator{\lwSt}{low\,St}

\DeclareMathOperator{\ind}{ind}
\usepackage[algo2e,ruled,vlined]{algorithm2e}

\usepackage{inputenc}
\usepackage{subcaption}

\newtheorem{lemma}{Lemma}
\usepackage{stmaryrd}
\usepackage{multirow}
\usepackage[table,xcdraw]{xcolor}
\newcommand{\ro}[1]{{\color{black}{#1}}}

\usepackage{multicol}
\usepackage{multirow}
\usepackage{mathrsfs}
\usepackage{amsmath,accents}
\usepackage{algpseudocode}
\usepackage{algorithm}
\usepackage{hyperref}

\usepackage{amsfonts}
\usepackage{pgf,tikz}
\usepackage{program}
\usepackage{natbib}
\usepackage{makecell, caption}
\usepackage{array}
    \newcolumntype{P}[1]{>{\centering\arraybackslash}p{#1}}
    \newcolumntype{M}[1]{>{\centering\arraybackslash}m{#1}}

\hypersetup{
pdftitle={Topology-based
Representative Datasets 
to Reduce Neural Network Training Resources},
pdfauthor={Rocio Gonzalez-Diaz et.al},
pdfkeywords={Data Reduction, Neural Networks, Representative Datasets, Computational Topology}
}

\begin{document}

\title{Emotion recognition in talking-face videos using persistent entropy and neural networks}


\author{ 
Eduardo Paluzo-Hidalgo\\
	Applied Math I department \\
    University of Sevilla\\
    epaluzo@us.es\thanks{Corresponding author.}\And
	\And
	Guillermo Aguirre-Carrazana\\
    University of Sevilla\\
    kaprekar.aguirre@gmail.com
	\AND
	Rocio Gonzalez-Diaz \\
    Applied Math I department \\
    University of Sevilla\\
    rogodi@us.es
}

\maketitle

\begin{abstract}
The automatic recognition of a person's emotional state has become a very active research field that involves scientists specialized in different areas such as artificial intelligence, computer vision or psychology, among others. Our main objective in this work is to develop a novel approach, using persistent entropy
and neural networks as main tools, to recognise and classify emotions from talking-face videos.Specifically, we  combine \ro{audio-signal and image-sequence}
information to compute a {\it topology signature} (a 9-dimensional vector) for each video. 
\ro{We  prove that small changes in the video produce small changes in the signature.}
These \ro{topological signatures} are used to feed a neural network to distinguish between the following emotions: neutral, calm, happy, sad, angry, fearful, disgust, and surprised.
The results reached  are promising and competitive, beating the performance reached in other state-of-the-art works found in the literature. 
\end{abstract}

\section{Introduction}
\ro{Facial emotion recognition consists of a series of processes to detect human emotions from facial human expressions.}
When people communicate with others, they are constantly sending and receiving nonverbal cues, expressed through body gestures, voice, facial expressions, and physiological changes. Nonverbal cues increase the trust, clarity and provide more information supporting what spoken words transmit. A particular emotional state produces certain verbal and nonverbal signals 
that transmit the information regarding personal feelings.

Nowadays, \ro{(facial)} emotion recognition has became an important  research area in the fields of computer vision and artificial intelligence due to its potential applications.
In general, people express their emotional state  (such as joy, sadness, or anger) through facial expressions and vocal tones and these are the features that are often  analized for emotion recognition.
So far, different approaches have been explored.
For example, in the H2020 KRISTINA project\footnote{http://kristina-project.eu/en/}, 
computer-aided emotion recognition is used to help in the interaction between health professionals and migrated patients, allowing to overcome linguistic barriers that 
hinder communication.
In addition to the KRISTINA project, other European projects working on emotion recognition are, for example, the   VocEmoApI project\footnote{ https://cordis.europa.eu/project/rcn/199804\_es.html} and the MixedEmotions project\footnote{https://cordis.europa.eu/project/rcn/194226\_es.html}. 
The VoicEmApI project developed a software for the detection of vocal emotion focused on extracting vocal markers that are caused by changes in physiological processes such as the cognitive affective process. Acoustic voice analysis works with the basic components of emotional processes, for example, a person's evaluation of relevant events or situations that could trigger actions and expressions that constitute an emotional episode. These evidences were continuously tracked.  The project deduced not only basic emotions, but also much finer distinctions, such as subcategories of emotions  and subtle emotions. The  project conducted  its impact to large markets such as home robotics, public safety, clinical diagnosis and therapy, call analysis and market research.
On the other hand, the MixedEmotions project developed an application based on a more complete emotional profile of the person's behavior. It  used data  from different channels: multilingual text data sources, audio and video signals, social media and structured data. The project offered commercial solutions providing an integrated big linked data platform for emotional analysis using heterogeneous sets of data and addressing the multilingual and multimodality aspects in a robust and large-scale setting.

Regarding  computer-aided emotion recognition research works, roughly speaking, they are   focused on the use of various input types such as facial expressions \cite{ofodile2017automatic,shojaeilangari2016pose, wan2017results}, speech \cite{plawiak2016hand, sapinski2018multimodal, kleinsmith2012affective, noroozi2018survey, avots2019audiovisual} and physical signals \cite{jenke2014feature}.

Furthermore, several classification and recognition techniques have been proposed in the past.
Some of them used speech prosody contours information to recognize emotions through different classification methods such as artificial neural networks, multi-channel hidden Markov models,  mixture of hidden Markov models, and, Active Appearance Models 
(AAMS).
Lately, hybrid neural networks combining convolutional neural networks  and recurrent neural networks  have become the state-of-art for emotion recognition.
For example,  the authors in \cite{guo2020audio} proposed an audiovisual-based hybrid network that combines the predictions of five models for emotion recognition in the wild. The overall accuracy of the proposed method achieved $55,61\%$ and $51,15\%$ classification accuracy for the audio-only and video-only dataset, respectively.
The authors used the Afew-va Database for Valence and Arousal Estimation In-the-wild  introduced in  \cite{kossaifi2017afew}.
The authors in \cite{issa2020speech} faced the task of  audio emotion recognition  using a convolutional neural network. Their baseline model included one-dimensional convolutional layers combined with dropout, batch normalization, and activation layer. 
The proposed framework achieved $71.61\%$ of accuracy
for the Ryerson Audio-Visual Database of Emotional Speech and Song (RAVDESS dataset) (see \cite{ravdess}).
 Later, in
 \cite{kwon2021mlt},  
 a new preprocessing scheme is proposed in order to remove the noise from speech signals based on fast Fourier transformation  and spectral analysis. 
The authors evaluated their model using benchmark IEMOCAP and EMODB datasets and obtained a high recognition accuracy, which were 73\% and 90\%, respectively. 
Finally,  in \cite{ctic2019},  a persistent-topology-based method  is developed to obtain a single value for the audio signal of a given video of a person expressing emotions. These data were later used as the input of a 
support vector  machine  to classify audio signals into eight different emotions, namely, neutral, calm, happy, sad, angry, fearful, disgust, and surprised. The results obtained were close to the existing accuracy of  methods with a greater scope such as the ones introduced in \cite{kryzhanovsky2017advances, zhang2015recognizing}.

This paper can be considered a continuation of the work \ro{developed} in \cite{ctic2019}. Here, instead of dealing with the audio signals \ro{only,} we deal with talking-face videos. 
The method incorporates the idea from \cite{ctic2019} for audio-signal emotion classification that consists of  computing the persistent entropy (introduced in \cite{PR2015}) of  the lower-star filtration of the 1-dimensional simplicial complex obtained by a discretization of the audio signal, together with the idea from \cite{gait1} for gait classification that consists of computing a 3-dimensional simplicial complex from the given image sequence, and extracting eight different topological  features from eight different filtrations considering, respectively, the distance to eight fixed planes (two horizontal, two vertical and four obliques).
\ro{This way, the method is able   to completely capture the movement in the image sequence. This methodology} were also   used   in \cite{javier2014} to monitor human activities at distance
and  in \cite{javier2014a} for gait-based gender classification.
Persistent entropy
has been widely used in very different situations. 
For example, in  \cite{chung2020persistence}, 
\ro{it} is used to measure the heart   rate   variability to a  sleep-wake classification. In \cite{PhysRevE.100.022314}, it is used
to detect dynamic states, and,  in \cite{matteo2020}, to detect glioblastomas.
Specifically, the persistence  barcode (or diagram) \ro{(a key tool in computational topology \cite{edelsbrunner2010computational})} is used to compute the persistence entropy  and, for the application considered in this paper, persistence barcodes have a clear interpretation  as the gestures will modify  the birth and death of the different connected components obtained from the facial landmark points through the considered filtration.
The novelty of the method presented here is the combination of the previous ideas with a particular construction of the 3-dimensional cell complex and the use of the result to feed a neural network.
Specifically, given a talking-face video, instead of segmenting each image of the image sequence as it is done in \cite{gait1}, we use precomputed landmark points to build the 2-dimensional  Delaunay triangulation in each frame  of the given image sequence and then stack  them in a particular way  to build a   3-dimensional cell complex. Later, we compute the persistent entropy  of  each of the eight filtrations.
The computed topological features 
\ro{together with the feature obtained from the audio-signal following \cite{ctic2019}, make up a 9-dimensional vector also called topological signature.
Besides, thanks to \cite{atienza2018stability}, we are able to prove that the topological signatures are stable to small changes in the input audio signal and image sequence.
Finally, the topological signatures computed}
are used  to train a neural network to classify emotions. 
Let us observe that the neural network considered in this paper is extremely simple because of the low dimension of the input.
This methodology has been tested using the   RAVDESS dataset \cite{ravdess} showing that  our results outperform  state-of-the-art methods.

The paper is structured as follows. The needed background is introduced in Section \ref{sec:background}. The description of the proposed method is provided in Section \ref{sec:method}. The experimentation made is presented in Section \ref{sec:experiments}, together with comparisons with state-of-the-art methods. Finally, Section \ref{sec:conclusion} provides conclusions and future work ideas.

\section{Background}\label{sec:background}

In this section,  the main concepts from topological data analysis  and neural networks, needed to understand  our method for facial emotion recognition, are recalled.
 
 \subsection{Topological data analysis}

Topological data analysis has emerged as an important approach to characterize the behavior of datasets using techniques from topology. Tools from topological data analysis, specifically persistent homology, allow assigning shape descriptors to large and noisy data across a range of spatial scales.  
Assuming that the input data are sampled from an underlying space $\X$ and that the aim is to recover the topology of $\X$, a general topological data analysis process follows these  steps: 
\begin{enumerate}
    \item Find an approximation of $\X$ using a combinatorial structure.
    \item Compute topological features of such structure such as persistent homology.
    \item Summarize the topological features computed using statistical tools such as persistent entropy.
\end{enumerate}

The combinatorial structure used in this work to represent objects is the one of  cell complexes, whose elements in each dimension $d$, called $d$-cells, are $d$-dimensional  topological spaces homeomorphic\footnote{A homeomorphism is a bicontinuous and bijective function between two topological spaces.} to a $d$-dimensional ball. This way, a 0-dimensional cell is a point (vertex), a 1-dimensional cell is  a curve, a 2-di\-men\-sion\-al cell is homeomorphic to a disk and so on. 
A {\it cell complex} $K$ is a collection of cells constructed inductively:
(1) The 0-skeleton $K^{(0)}$  (i.e., the set of 0-cells of $K$) are a set of points in an ambient $n$-dimensional space $\mathbb R^n$.
(2) Form the $d$-skeleton  $K^{(d)}$ from the $(d-1)$-skeleton $K^{(d-1)}$ by attaching $d$-cells via homeomorphisms.

\begin{figure}[ht!]
\begin{center}
\includegraphics[width=\linewidth]{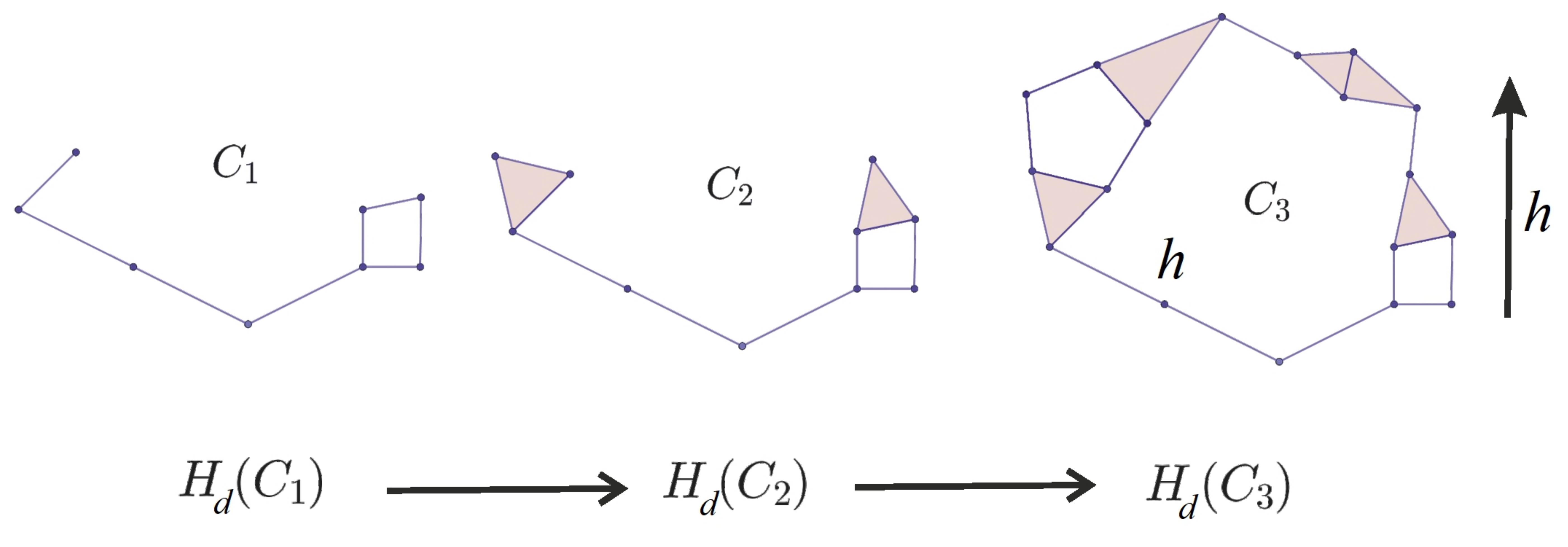}
\end{center}
\caption{Top: Example of a filtration obtained using the height function $h$ on its vertices. Bottom: Associated $d$-dimensional persistent homology.
{For example, if $d=1$ then 
$H_1(C_1)=\mathbb{Z}_2\stackrel{[1]}\longrightarrow H_1(C_2)=\mathbb{Z}_2\stackrel{[1\;0]}\longrightarrow H_1(C_3)=\mathbb{Z}_2\oplus \mathbb{Z}_2$ }
}
\label{fig:1}
\end{figure}

From now on, we will assume that the given cell complex $K$ has a finite number of cells. 
The  boundary {set} of a $d$-cell $\sigma\in K$ can be informally defined as the {set of} $(d-1)$-cells in the $(d-1)$-skeleton $K^{(d-1)}$ used to attach the $d$-cell $\sigma$. {Successively adding to $F=\{\sigma\}$ the boundary set  of each  cell in $F$, we obtain the faces of $\sigma$}.
For example, the boundary {set} of an edge is its two endpoints (vertices). 
A maximal cell of $K$ is a cell     not in the boundary of any other cell of $K$. A $d$-dimensional cell complex $K$ is     a cell complex satisfying that the dimension of the cell of the higher dimension in $K$ is $d$. 
A  subcomplex of a cell complex $K$ is a subset $K' \subset K$ which itself is still a cell complex.
An example of a subcomplex is the closed star of a vertex $v$, denoted by $\St\, v$ and defined as follows: A cell $\sigma$ is in $\St\, v$ if   there exists $\mu\in K$ such that $\sigma$ and $v$ are faces of $\mu$.
A filtration is an increasing sequence of cell complexes $\emptyset\subseteq K_1 \subset K_2 \subset \cdots \subset K_r= K.$
A particular case of cell complexes is the one of simplicial complexes. For any dimension $d$, 
a $d$-simplex is the convex hull of $d+1$ affinely independents points $v_0, v_1, \cdots, v_{d} $ living in the $n$-dimensional space $\mathbb{R}^n$ with $d\leq n$. 
See an example of a filtration of simplicial complexes in Figure \ref{fig:1}.

There are several methods to compute cell complexes and filtrations from input data depending on the nature of the data and the purpose of the analysis. In this work, we will use the following process to produce simplicial complexes  from data: 
Consider a  set of points $S$ in $\mathbb{R}^n$.
Define $V_s$ as the set of points of $\mathbb{R}^n$ that are closer to $s\in S$ than to any other point of $S$. That is, for $s\in S$, $V_s =  \{ x \in \mathbb{R}^{n} \ \mid \; d(x,s) \leq d(x, s') \ \forall s' \in S\}. $
The collection of the sets $V_s$ is a covering for $\mathbb{R}^n$ and it is called the Voronoi decomposition of $\mathbb{R}^n$ concerning $S$.
The nerve of this covering is a a simplicial complex called the Delaunay triangulation of $S$.
The construction of this complex is costly in high dimensions, although there exist efficient algorithms for computing it  when $n=2$ and $n=3$. See \cite{toth2017handbook} for more details on  Voronoi diagrams and  Delaunay triangulations. 
The filtration considered in this paper is the lower-star filtration. Let us see how to define it. 
Consider a real-valued function  $h$ on  a finite set of points $V\subset \mathbb{R}^n$. Suppose  $K$  is a cell complex with set of  vertices $V$.
The  lower star of $v\in V$  is defined as the subset of
cells of $K$ for which $v$ is the vertex with maximum function value, that is,
$\lwSt \,v = \{\sigma\in \St\,v: x\in \sigma \Rightarrow h(x) \leq h(v)\}.$
Sort the vertices  by their function values, in a non-decreasing order,  $V=\{v_1, v_2, \dots, v_r\}$. 
The   lower-star filtration 
(see \cite[page 135]{edelsbrunner2010computational})
$\emptyset\subseteq K_1 \subset K_2 \subset \cdots \subset K_r= K$
 satisfies that $K_j$ is  the union of the lower stars of the first $j$ vertices of $V$, that is, $K_j=\bigcup_{i\leq j} \lwSt \,v_i$, for all $j$. 
Once we have computed a filtration, the next step is to compute topological features from it, such as the  homology.
The $d$-dimensional homology group of a topological space represents its $d$-di\-men\-sion\-al holes. Intuitively, a 0-dimensional hole is a connected component, a 1-dimensional hole is a tunnel and a 2-dimensional hole is a cavity. 
Classical results ensure that $d$-dimensional homology groups are topological invariant, that is, they are invariant under homeomorphisms.
The $d$-di\-men\-sion\-al homology group of a topological space structured as a cell complex $K$ can be defined as follows. 
First,  the $d$-dimensional chain group $C_d(K)$ is obtained by summing up $d$-cells\footnote{The ground ring considered in this paper is $\mathbb{Z}/\mathbb{Z}2$.}. Then, the boundary operator is extended to a linear map $\partial_d$ from $C_d(K)$ to $C_{d-1}(K)$ in the obvious way. Since the boundary of the boundary of a cell is always zero, then the image $B_d(K)$ of $\partial_{d+1}$ is a subgroup of the kernel $Z_d(K)$ of $\partial_d$ and then, the $d$-dimensional homology group of $K$ is the quotient group $H_d(K)=B_d(K)/Z_d(K)$. An element $\alpha$ of $H_d(K)$ is called a $d$-dimensional homology class of $K$.

Although the concept of homology is not useful in practice due to its lack of discrimination,  this notion is used in \cite{edelsbrunner2010computational}
to define a more discriminant concept called persistent homology, together with an efficient algorithm to compute it. Later,   in \cite{zomorodian2005computing},  the initial definition is reformulated.
The persistent homology of a filtration  tracks  the moment $i$  where a homology class is born and the moment $j$ where the same class  dies leading to a topological descriptor called persistence  diagram.  
For example the lower-star filtration 
leads a family $\{ H_d(K_t): t \in \mathbb{R}\}$ of homology groups and the inclusions $K_t \hookrightarrow K_s$ lead a family of homomorphisms $\{ H_d(K_t) \rightarrow H_d(K_s):
t\leq s\}$,  {for $d\in\mathbb{Z}$.}
Now, each   homology class $\alpha$ that was born in $H_d(K_t)$ and died in $H_d(K_s)$ can be stored as a point $(t,s)$. The result is a multi-set of points in $\mathbb{R}^2$ called the persistence diagram for the given filtration.
The persistence of the homology class $\alpha$  is the difference $\pers(\alpha) = s-t$. Then,  homology classes with infinity persistence correspond  to points of the form $(t, +\infty)$.
 In this work, 
 points of the form $(t, +\infty)$ are replaced by points of the form $(t,N+1)$,
 where $N$ is a fixed big positive integer. This way,  all points in the persistence diagram have finite coordinates. 
The features of higher persistence are represented by the points furthest from the diagonal while nearby points to the diagonal may be interpreted as topological noise. 

 In \cite{turner2014frechet,munch2015probabilistic,chazal2013bootstrap}
 among others,  persistence diagrams are studied from a probabilistic and statistical point of view. 
In this paper, 
we  summarize the information described by a persistence diagram in a number called persistent entropy (introduced in \cite{PR2015})  which consists in the Shannon entropy of the probability distribution obtained from the given persistence diagram. There are many applications of persistent entropy. For example, in  pattern recognition \cite{merelli2016topological, rucco2017new}, complex systems  \cite{binchi2014jholes,PhysRevE.100.022314}, 
time series \cite{chung2020persistence}, and clustering \cite{wang2017scale}. 
Given a filtration  and the corresponding persistence diagram $\Dgm=\{(a_j, b_j): j \in J\}$,  
the persistent entropy of the filtration is defined as
$ E= - \sum_{j \in J} p_j \, \log(p_j)$
where $p_j = \frac{\ell_j}{L}$, $\ell_j = b_j - a_j$, and $L= \sum_{j \in J} \ell_j$. 
Let us notice that if $p_j\leq 1$, then $log(p_j) \leq 0$, so the persistent entropy  is always positive.
Intuitively, the persistent entropy measures how different the persistence  of the homology classes that appear along the filtration are.

\begin{figure}
    \centering
    \includegraphics[width=0.45\textwidth]{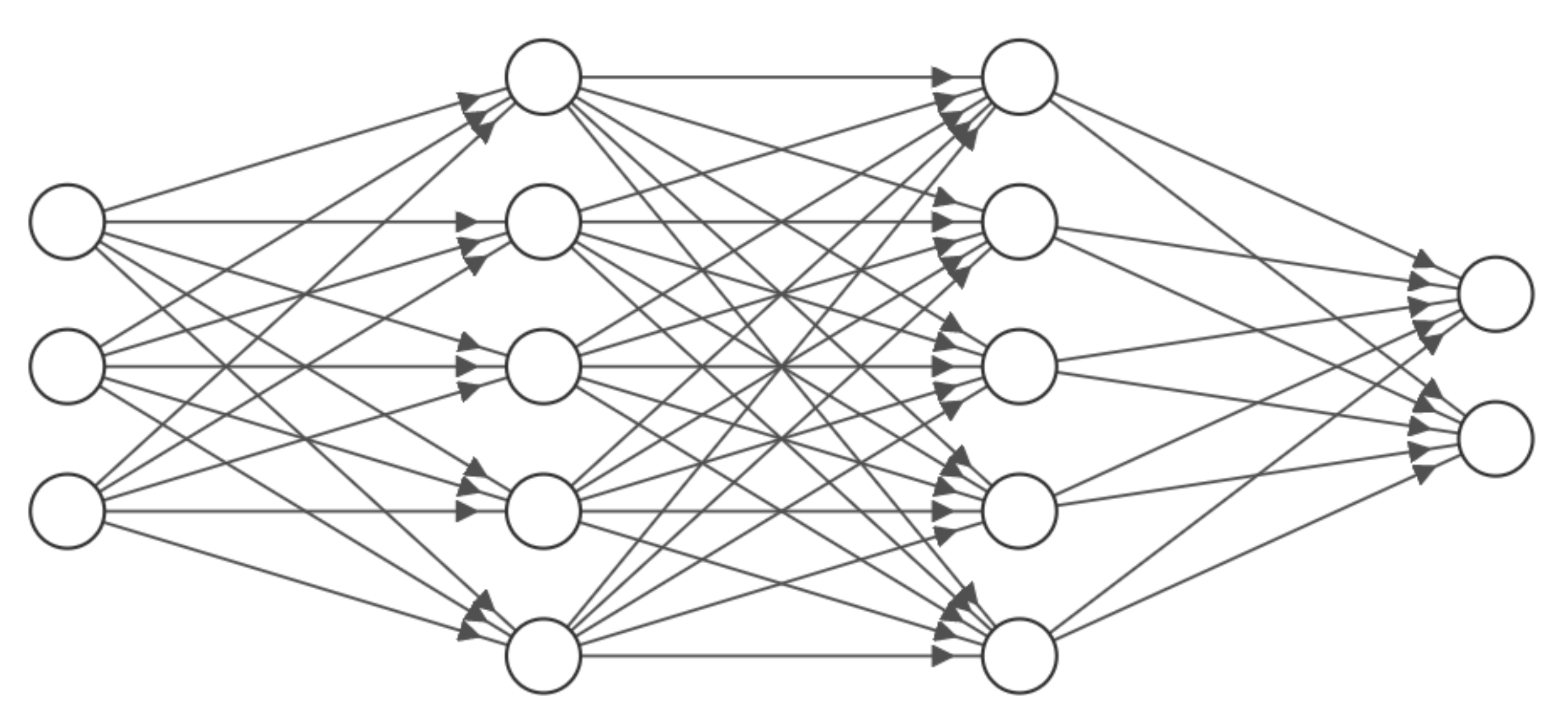}
    \caption{A $3\times 5\times 5\times 2$ feedforward neural network composed of an input layer with 3 neurons, two hidden layers with 5 neurons each, and an output layer with 2 neurons.}
    \label{fig:generic_nn}
\end{figure}

\subsection{Neural Networks}

In this paper, we deal with a supervised classification problem where a set of labelled examples are provided with the aim of making predictions for unlabelled points. A widely extended machine learning model for classification problems is neural networks.
In general, 
we could say that a neural network is a mapping ${\cal N}_{\omega,\Phi}: {\mathbb R}^n \to {\mathbb R}^m$ that depends on a set of weights
$\omega$ and a set of parameters $\Phi$ describing the synapses between neurons, layers, activation functions and any other characteristic of its architecture.
 A good introduction to artificial neural networks was given in \cite{haykin2009neural}.
A specific kind of neural network architecture is a feedforward neural network  composed of a set of neurons hierarchically organized in layers that are fully connected as the example displayed in  Figure~\ref{fig:generic_nn}. Neural networks can be seen as directed graphs where the input is transmitted and transformed along the graph using different operations. In each neuron (represented as a node  of the graph), an activation function such as ReLu, sigmoid or softmax  is applied.

To train the neural network ${\cal N}_{w,\Phi}$ for a supervised classification task,
we will use a labelled dataset $D= \{(x,c_x)\}$ consisting of a finite set of pairs  where, for each pair $(x,c_x)$, point $x$ lies in ${\mathbb R}^n$ and label $c_x$ lies in $\{0,1,\dots,k\}$, for some $k\in\mathbb{N}$.
During the training process, 
the set of weights of the neural network is updated trying to minimize a loss function  which measures the difference between the output of the network (obtained with the current weights) and the desired output (dictated by the labelled dataset). 
The loss function used in this paper is the cross-entropy loss function which is related to the Kullback-Leibler divergence:
Given two probability distributions $P(x)$ and $Q(x)$ over the same random variable $x$, the cross-entropy is computed as $H(P,Q)=-\sum_{(x,c_x)\in D}P(x)\log(Q(x))$. 
To iteratively update the weights, the loss-driven  training method  used in this paper   is the Adam algorithm (introduced in \cite{kingma2017adam}) which is a stochastic gradient-based optimization algorithm.

The goal  of training a neural network is generalization. That is, we want our neural network to learn from the given data and to apply the learnt information on new data.
One way to measure the performance of the trained neural network is to split the given dataset into two subsets called the training set and the test set.
When the trained neural network reaches high accuracy on the training set but performs badly on new data, we say that there is an overfitting. There are different approaches to prevent overfitting such as dropout that consists in invalidating randomly a certain percentage of the neurons of the neural network during the training procedure (consult  \cite{10.5555/2627435.2670313} for more information). 

\begin{figure}[ht!]
\begin{center}
\includegraphics[width = 0.5\linewidth]{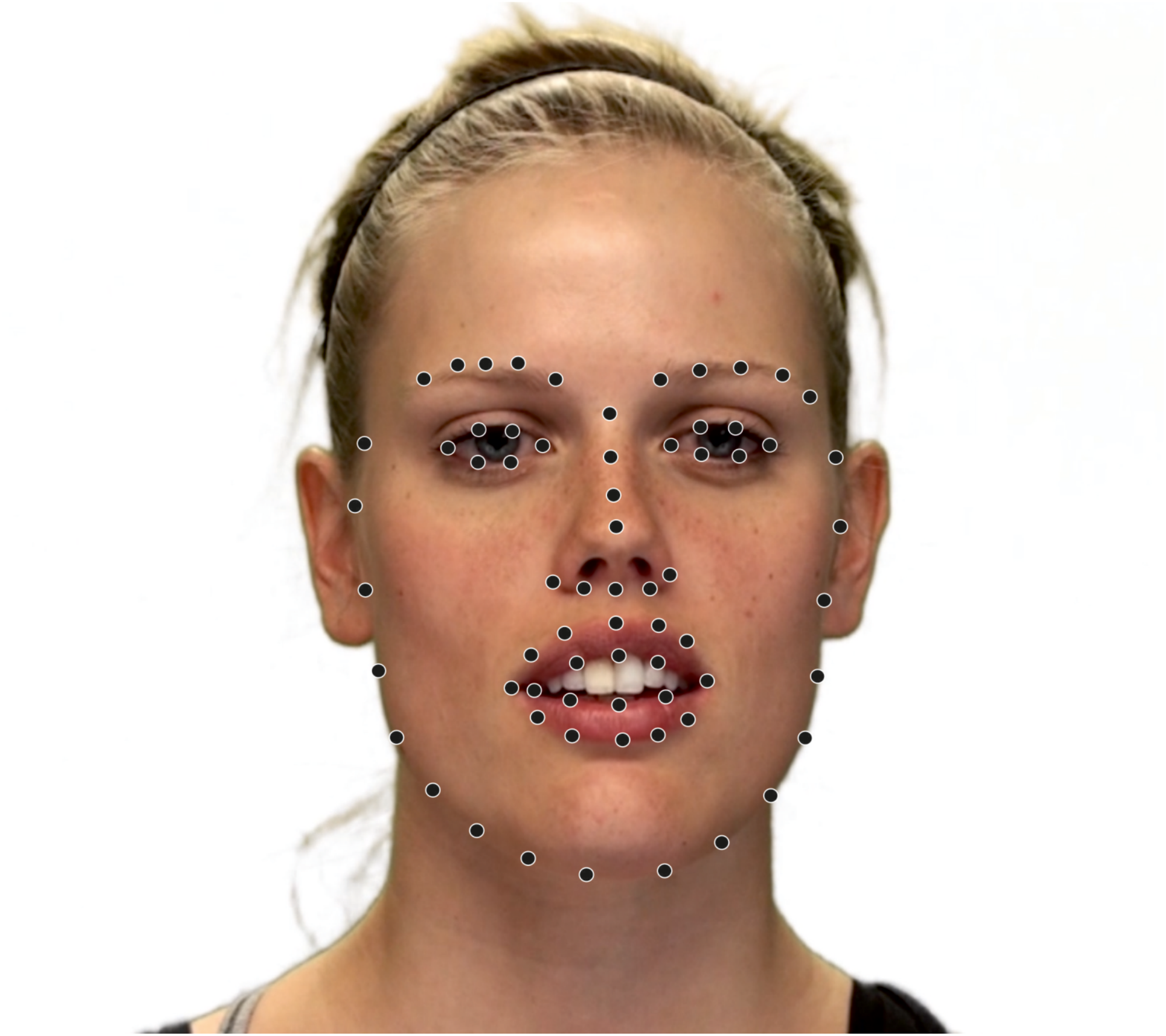}
\caption{The  landmark points considered in this paper, drawn on a face in one frame of a video from the RAVDESS dataset.
}
\label{fig:landmarks}
\end{center}
\end{figure}

\section{Description of the method}\label{sec:method}

 In this section, we develop an  emotion recognition method using  persistent entropy and neural networks as main tools. 
 Overall, the method works as follows. 
 The input data  are \ro{talking-face videos} 
 with precomputed facial landmark points.
 For each video, 
 \ro{we compute a topological feature  obtained from the audio signal together with eight topological features obtained from the image sequence, deriving a 9-dimensional vector call the topological signature of the video.} 
 The set of topological signatures obtained from the video dataset will then be used to feed a neural network.


We start the procedure by extracting the
landmark points on each frame of the input image sequence (see Figure \ref{fig:landmarks}). \ro{we assume that these landmark points are already precomputed in the given dataset.}

\begin{figure}[ht!]
\begin{center}
\includegraphics[width =0.45  \linewidth]{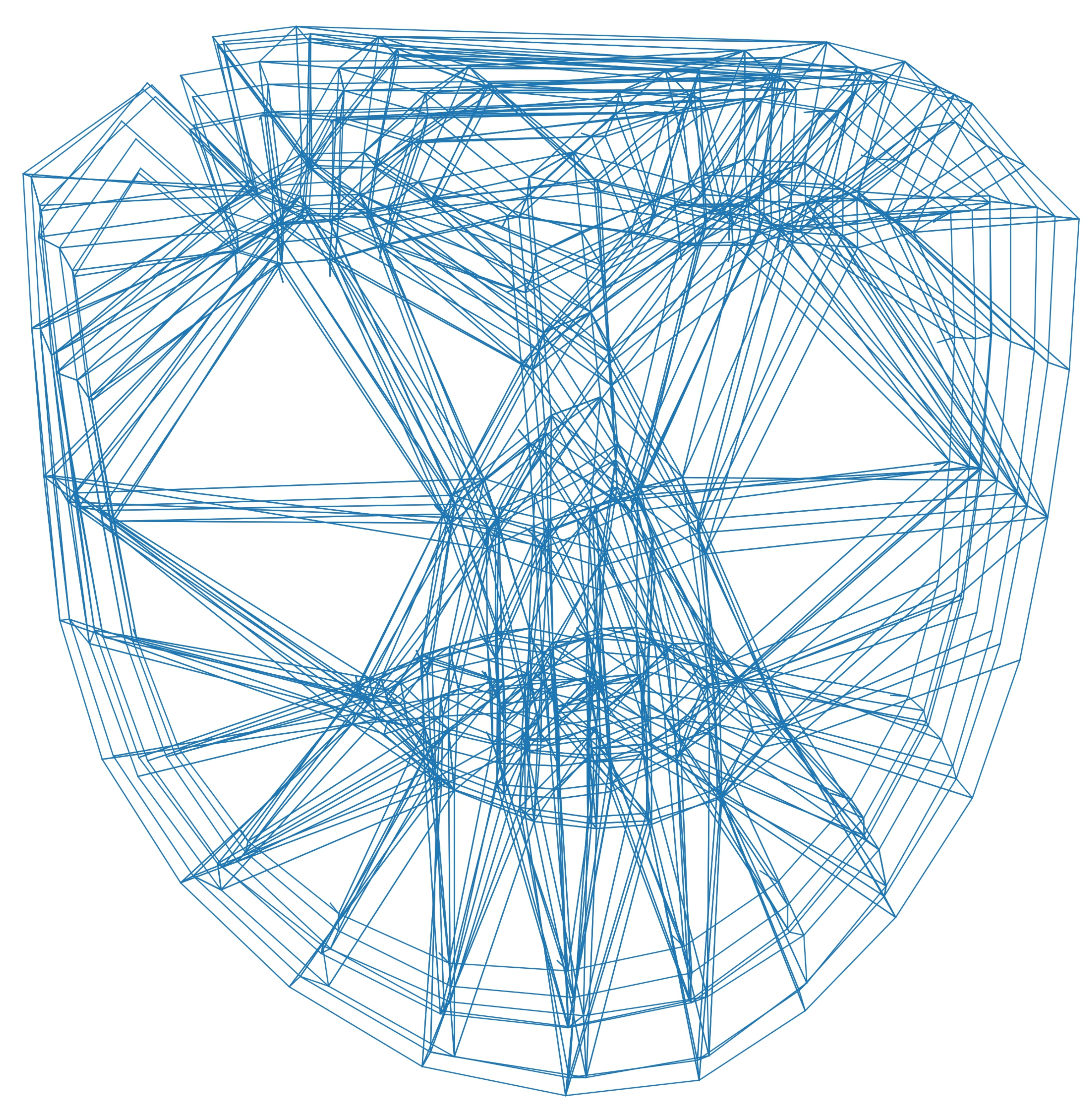}
\caption{The 1-skeleton of the cell complex  obtained from an image sequence.}
\label{fig:complex}
\end{center}
\end{figure}

Given an image sequence obtained from a talking-face video \ro{with landmark points precomputed,} we compute,
for each frame, 
the  2-dimensional simplicial complex consisting of the Delaunay triangulation of the set of points corresponding to the spatial position of the landmark points. 
In order to connect the topological information along the image sequence, 
the landmark points corresponding to the same part of the face in consecutive frames are joined by an edge.
A 2-dimensional cell is obtained when the two endpoints of an edge are joined to the two endpoints of the corresponding edge in the neighbor frame. 
A 3-dimensional cell is obtained when the vertices of a triangle \ro{of the Delauney triangulation associated to one frame} are joined with the vertices of the corresponding triangle in the neighbor frame. 
The output of the steps described in this section is a 3-dimensional cell complex $K$ for each input \ro{image sequence} which condenses all the gestures the person is making while recording on video. 
See Figure \ref{fig:complex}, 
where the 1-skeleton  of the 3-dimensional cell complex $K$ obtained from an image sequence is pictured. 

\begin{figure}
    \centering
    \includegraphics[width = 0.7\linewidth]{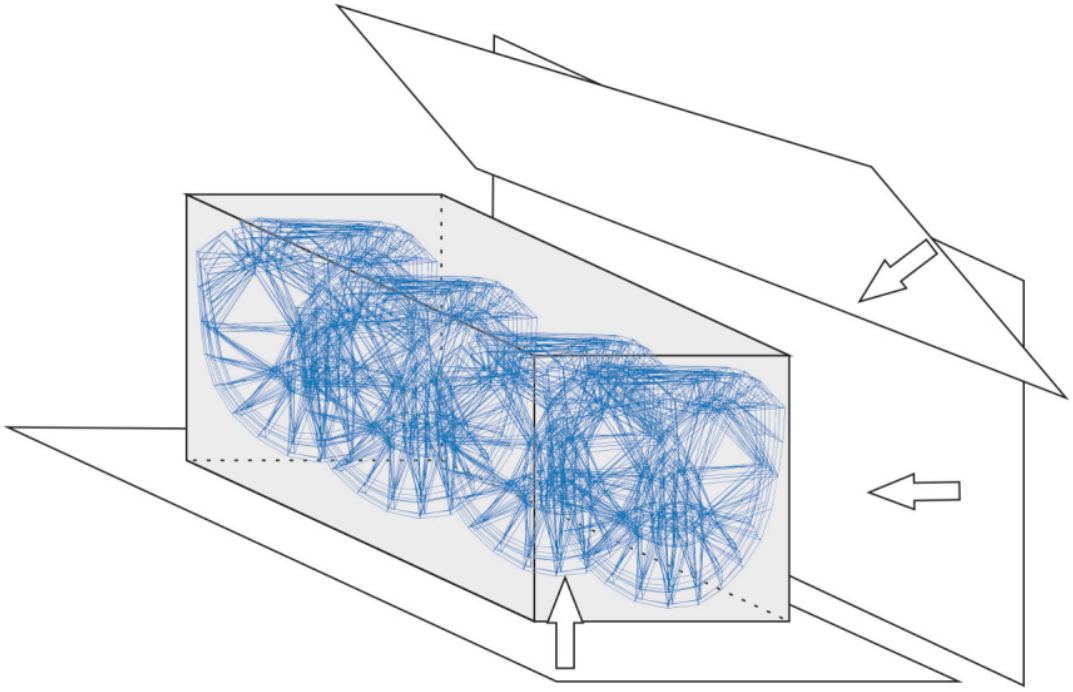}
    \caption{Illustration of three of the eight different filtrations considered.}
    \label{fig:example_filtration}
\end{figure}

The next step in this process is to sort the cells of $K$ in order to obtain a filtration. 
In this work, 
eight different filtrations (two horizontal, two vertical, and four obliques) are used to obtain eight different persistence diagrams (see Figure~\ref{fig:example_filtration} to have intuitions). 
The filtrations will capture the small movements of the landmark points through the given image sequence. 
The way to define a filtration is as follows: 
Given a plane $\pi$, 
we define the filter function $h_{\pi}:  K \rightarrow \mathbb{R}$ that assigns to each  vertex of $K$ its distance to the plane $\pi$, and to any other cell of $K$,
the maximum distance of its vertices to the plane $\pi$.
The cells are sorted according to the function values of their vertices, and then, the lower-star filtration $ K_{\pi}$ associated with the plane $\pi$ is computed.

Next, the persistence diagram is computed for each of the eight filtrations. 
The algorithm used for this step  is described in Algorithm \ref{alg} {with complexity $O(n^3)$ in theory but linear in practice \cite[p.~159]{edelsbrunner2010computational}}.

The persistent entropy is then computed for each of the persistence diagrams obtained. Due to its formulation, persistent entropy can be computed in linear time.
As a result, 
an 8-dimensional vector is obtained for each image sequence.

\begin{algorithm2e}
  \KwIn {A filtration $\emptyset= K_0 \subset K_1\subset K_2\subset \dots \subset K_n=K $ and an ordering of the cells 
  $\{\sigma_1,\dots,\sigma_m\}$
 of $K$ 
 such that if $i<j$ then $\ind(\sigma_i)<\ind(\sigma_j)$ where $\ind(\sigma_i)=\min \{r\,:\, \sigma_i\in K_r\}$}
  \KwOut{The persistence diagram  $\Dgm$}
Initialize  $H=\emptyset$,
 $\Dgm=\emptyset$, and $f(\sigma_i)=0$  for $i\in\{1,\dots,m\}$\\
\For{$i=1$ {\bf to} $m$}
	{\If{$f\partial (\sigma_i)==0$}
    		{	
    		$ H\cup\{\sigma_i\}$
    		\mbox{     } (a new homology class was born)\\
    		$f(\sigma_i)=\sigma_i$\\
    		$ \Dgm\cup\{(\ind(\sigma_i),\infty)\}$
		}
\If{$f\partial (\sigma_i)\neq 0$}
		{Let $\sigma_j\in f\partial (\sigma_i)$ such that $ j==\max\{\ind(\mu): \mu\in f\partial(\sigma_i)\}$\\
			$H\setminus\{\sigma_j\}$
			\mbox{     }
			(an homology class died) \\
		\ForEach{{$x\in K$} such that $\sigma_j\in f(x)$}
			{$f(x)= f(x)+f\partial(\sigma_i)$.}	
			$\Dgm\setminus\{(\ind(\sigma_j),\infty)\}\cup 
			\{(\ind(\sigma_j),\ind(\sigma_i))\}$
			}
	}
  \caption{Computing the persistence diagram
for  a filtration   \cite{at-model}.}
  \label{alg}
\end{algorithm2e}

\begin{figure*}
    \centering
    \includegraphics[width=\textwidth]{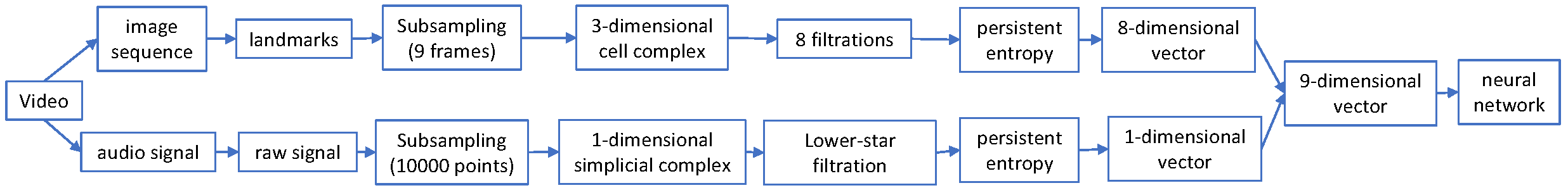}
    \caption{Summary of the workflow for emotion classification used in this paper.}
    \label{fig:workflow}
\end{figure*}

Finally, 
for each talking-face video, 
we add a new entry to the  8-dimensional vector 
computed
consisting of the persistent entropy of the  lower-star filtration obtained from the 1-dimensional simplicial complex computed from the raw audio signal of the video as it is done in \cite{ctic2019}.

\ro{Putting all together, we obtain} a 9-dimensional feature vector called the {\it topological signature} of the video.
Moreover, thanks to the work presented in \cite{atienza2018stability} we have the following result.
\begin{lemma}
\ro{The so-called topological signature associated to a given talking-face video is} stable in the sense that  small changes in the video produce small changes in the signature. 
\end{lemma}
\noindent{\bf Proof.}
In \cite{atienza2018stability}, it is proved that  persistent entropy is stable. It means that small changes in the input data produce small changes in the persistent entropy value. In this case, input data are, first, the eight filtrations obtained from the image sequence and, second the filtration obtained from the audio signal. Finally, small perturbations in the filtrations are equivalent to a small displacement of the landmark points in the image sequence or small changes in the audio signal, that is, they consists of small perturbation in the input data used to compute the persistent entropy values, concluding the proof. 
\hfill{$\Box$}

The topological signatures computed from the talking-face video dataset are then used to train a feed-forward neural network to classify the videos  into the different emotions considered.

\begin{table*}[hbt!]
\caption{Comparison of our method with state-of-the-art methods.}
\label{comparison}
\centering
\begin{tabular}{|l|l|c|}
\hline
Paper    
& Dataset & Average accuracy \ro{on the test set} \\ \hline
   \cite{sungwoo2020}                                          & RAVDESS   & 87.11\%  \\ \hline
\cite{mti4030046} & RAVDESS  & 86.36\%  \\ \hline 
\cite{MA2019184}  & RML, enterface05, AUM-1s  & 77.31\%\\ \hline
\cite{WANG2020115831} & RAVDESS  &77.66\% \\
\hline
Our method  & RAVDESS      & \textbf{95.97\%}   \\ \hline
\end{tabular}
\end{table*}

\section{Experimentation}\label{sec:experiments}

For experimentation, 
the  Ryerson Audio-Visual Database of Emotional Speech and Song (RAVDESS) is used,
\ro{that is a  talking-face video dataset 
where facial landmark points  
composed of 62 points have been precomputed.}
This dataset contains the vocalization of two statements in a neutral North American accent by 24 professional actors (12 female, 12 male). 
Each expression is produced at two levels of emotional intensity (normal, strong), with an additional neutral expression.
The intensity fulfils an important role in emotional theory (see the works in \cite{diener1985intensity,schlosberg1954three}). 
The strong intensity is useful when we are looking for clear emotional examples. However, as explained in \cite{kaminska2014recognition}, 
the normal intensity is generally used if we are interested in providing classification for daily life. 
All actors produced  60 spoken expressions and 44 sung expressions. 
These vocalizations are available in three formats: audio-only, video-only, and audio-video.
Besides, the RAVDESS video dataset contains tracked facial landmark points for all videos.

In this paper, 
we focus on the 60 speech videos provided in the RAVDESS video dataset. 
The total tracked files used is 24 actors $\times$ 60 speeches. 
Since we do not consider neutral emotions to avoid an unbalanced dataset, 
we used a total of 1344 videos.
The number of frames used 
\ro{as well as the number of points in the subsampled audio signal were}
a experimental choice consisting of the minimum number of frames \ro{and points} needed to obtain good results and to develop the experiment in a feasible time. 
The neural network considered was the simplest one that provided satisfactory results and the weights of the neural network were tuned using a traditional training procedure.


\begin{table*}[hbt!]
\caption{Confusion matrix of the audio-video experiment for one of the repetitions measured on the test dataset.}\label{confusion_mat}
\centering
\begin{tabular}{|l|c|c|c|c|c|c|c|}
\hline
\multicolumn{1}{|c|}{Emotion} & Calm                       & Happy                      & Sad                        & Angry                      & Fearful                    & Disgust                    & Surprised                  \\ \hline
Calm                          & \cellcolor[HTML]{DAE8FC}61 & 0                          & 0                          & 0                          & 0                          & 0                          & \cellcolor[HTML]{FCFF2F}3  \\ \hline
Happy                         & 0                          & \cellcolor[HTML]{DAE8FC}48 & 0                          & \cellcolor[HTML]{FCFF2F}4  & 0                          & 0                          & 0                          \\ \hline
Sad                           & 0                          & 0                          & \cellcolor[HTML]{DAE8FC}55 & 0                          & 0                          & 0                          & 0                          \\ \hline
Angry                         & 0                          & 0                          & 0                          & \cellcolor[HTML]{DAE8FC}60 & 0                          & 0                          & \cellcolor[HTML]{FCFF2F}2  \\ \hline
Fearful                       & 0                          & 0                          & 0                          & 0                          & \cellcolor[HTML]{DAE8FC}60 & 0                          & \cellcolor[HTML]{FCFF2F}2  \\ \hline
Disgust                       & 0                          & 0                          & 0                          & 0                          & 0                          & \cellcolor[HTML]{DAE8FC}45 & \cellcolor[HTML]{FCFF2F}1                          \\ \hline
Surprised                     & 0                          & 0                          & 0                          & 0                          & \cellcolor[HTML]{FCFF2F}4  & 0                          & \cellcolor[HTML]{DAE8FC}55 \\ \hline
\end{tabular}
\end{table*}

The steps for the experimentation were the following (see Figure \ref{fig:workflow}). 
We first consider the image sequence obtained for each video of the audio-video dataset.
For each image sequence, 9 equally spaced  frames were selected  to have an appropriate representation of the full image sequence. 
The landmark points of those 9 frames were then used to build an 8-dimensional vector following the method described in Section~\ref{sec:method}.
Then,  
for each video, 
we added a new entry to the 8-dimensional vector computed consisting of the 
persistent entropy of the  lower-star filtration 
\ro{of the 1-dimensional simplicial complex obtained from  a subsampling of the raw audio signal consisting of 
10000 points. The subsampling process was
done uniformly on the signal, maintaining its shape and main distribution of the spikes.}
As a result, 
we obtained a set of 1344 9-dimensional feature vectors, one for each video of the dataset considered.
 
 Finally, this set  was split into a training set with 944  vectors and a test set for validation with  400  vectors. 
  Then, 
  the training set was used to train a neural network with the following standard architecture: 
It is composed by 5 layers with a total of
 $n\times 512\times 128\times 64 \times 7$ neurons, using dropout ($20\%$) in the first hidden layer with $n=9$ being the dimension of the input 
 (i.e., the 9-dimensional topological feature vectors). 
 The hidden layers are composed by  the ReLU activation function.
 Finally, the Softmax activation function is put in the output layer.
 
\begin{figure}
    \centering
    \includegraphics[width=0.45\textwidth]{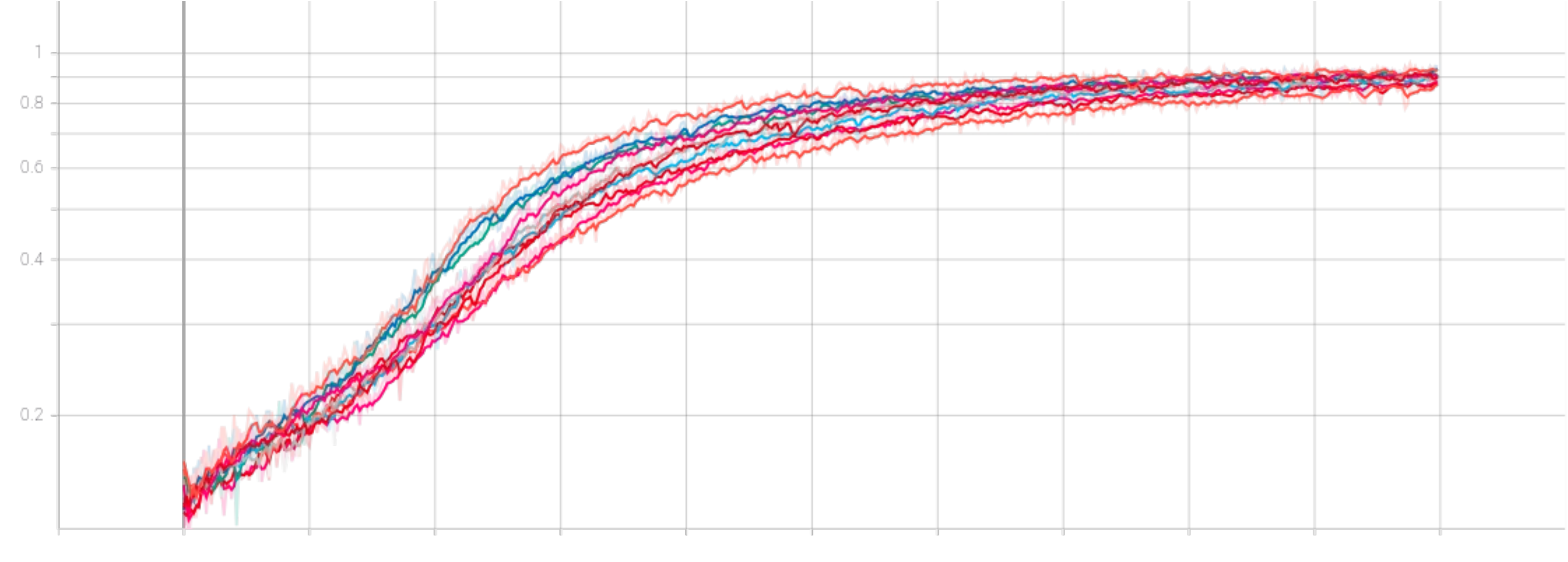}
    \caption{Accuracy values on the training set during 500 epochs. 10 repetitions.}
    \label{fig:accuracy_train}
\end{figure}

\begin{figure}
    \centering
    \includegraphics[width=0.45\textwidth]{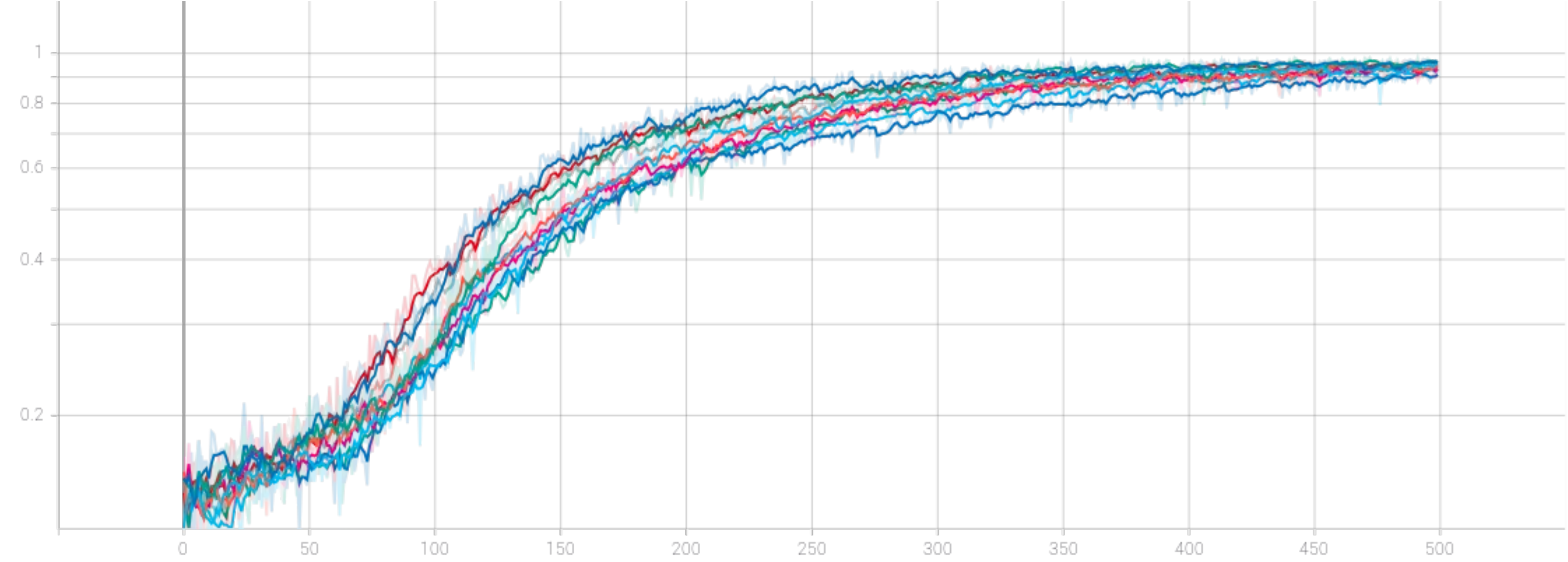} 
    \caption{Accuracy values on the test dataset during 500 epochs. 10 repetitions.}
    \label{fig:accuracy_test}
\end{figure}

 The neural network was trained during 500 epochs and the experiment was repeated 10 times using sparse categorical cross-entropy as loss function and the Adam training algorithm. 
 The accuracy values for those repetitions are shown in Figure~\ref{fig:accuracy_train} for the training set and in Figure~\ref{fig:accuracy_test} for the test set. 
 
The highest values reached were  $99.9\%$ of accuracy  on the training set with $98.02\%$ of accuracy on the test set. \ro{Average accuracy was 95.97\% on the test set.
A confusion matrix for the experiment is shown in Table~\ref{confusion_mat}.}

The state-of-the-art methods with which we compare ours are the following ones. 
In \cite{sungwoo2020}, 
a model is proposed based on three deep networks that are fed using image sequences, facial landmark points and acoustic features, respectively. 
In \cite{mti4030046},  
the fusion of visible images and infrared images with speech are used  to feed an ensemble method based on convolutional neural networks.  
The authors in \cite{MA2019184} applied a convolutional neural network approach  with a prepocessing method to eliminate data redundancy and noise. 
Finally, 
the authors in \cite{WANG2020115831} used  convolutional and recurrent neural network together with long short-term memory.
As we can see in Table~\ref{comparison}, 
our method outperforms all of them.

 The code developed is available at the link \url{https://github.com/Cimagroup/AudioVisual-EmotionRecognitionUsingTDA}.
 All the parameters are provided in the implementation and the replication of the experiments on the RAVDESS database can be done using the code without problem. 
If other dataset different to RAVDESS is used then the facial landmark points should be computed before applying the algorithm proposed in this paper. 

\section{Conclusions and future work}\label{sec:conclusion}

In this work, we have developed a novel method using  persistent entropy and neural networks \ro{for emotion classification of talking-face videos.}
The results reached  are promising and competitive, beating the performance reached in other state-of-the-art works found in the the  literature. 
We  combined audio-signal and image-sequence information to develop our topology-based emotion recognition method.
The main drawback of our methodology is the need of a  video long enough to be able to select a representative subset of frames to compute the cell complex. 
This fact makes our method not useful in real-time applications.

The following future works are plan to be explored: 
To expand the topological signature by extracting more information from the audio signals. 
To divide the landmark points into different subsets to determine regions or pairs of regions that contain discriminative landmark points for each facial expression. 
To use the the 3-dimensional information provided by the landmark points. 
To take advantage of the depth information of the landmark points could be a challenging problem for the future together with considering higher dimensional topological information once that we increase the dimension of the data we are dealing with.

\bibliographystyle{plainnat}
\bibliography{biblio}


\end{document}